\documentclass[lettersize,journal]{IEEEtran}
\usepackage{amsmath,amsfonts}
\usepackage{algorithmic}
\usepackage{algorithm}
\usepackage{array}
\usepackage[caption=false,font=normalsize,labelfont=sf,textfont=sf]{subfig}
\usepackage{textcomp}
\usepackage{stfloats}
\usepackage{url}
\usepackage{verbatim}
\usepackage{graphicx}
\usepackage{cite}
\usepackage{multirow}
\usepackage[table,xcdraw]{xcolor}
\usepackage{colortbl}

\begin{document}

\title{Multi-task Feature Enhancement Network for No-Reference Image Quality Assessment}

\author{Li Yu

}

\markboth{Journal of \LaTeX\ Class Files,~Vol.~14, No.~8, August~2021}%
{Shell \MakeLowercase{\textit{et al.}}: A Sample Article Using IEEEtran.cls for IEEE Journals}

\IEEEpubid{0000--0000/00\$00.00~\copyright~2021 IEEE}

\maketitle

\begin{abstract}
Due to the scarcity of labeled samples in Image Quality Assessment (IQA) datasets, numerous recent studies have proposed multi-task based strategies, which explore feature information from other tasks or domains to boost the IQA task. Nevertheless, multi-task strategies based No-Reference Image Quality Assessment (NR-IQA) methods encounter several challenges. First, existing methods have not explicitly exploited texture details, which significantly influence the image quality. Second, multi-task methods conventionally integrate features through simple operations such as addition or concatenation, thereby diminishing the network's capacity to accurately represent distorted features. To tackle these challenges, we introduce a novel multi-task NR-IQA framework. Our framework consists of three key components: a high-frequency extraction network, a quality estimation network, and a distortion-aware network. The high-frequency extraction network is designed to guide the model's focus towards high-frequency information, which is highly related to the texture details. Meanwhile, the distortion-aware network extracts distortion-related features to distinguish different distortion types. To effectively integrate features from different tasks, a feature fusion module is developed based on an attention mechanism. Empirical results from five standard IQA databases confirm that our method not only achieves high performance but also exhibits robust generalization ability.
\end{abstract}


\section{Introduction}
\IEEEPARstart{I}{n} recent years, with the advent of the digital age and the rapid development of technology, image data has shown explosive growth. However, the diversity of images and equipment has caused the problem of uneven image quality. At the same time, the image will be affected by a variety of factors in the process of acquisition, transmission, etc., which leads to the distortion and quality change of the image. This ultimately affects people's perception of images and the effectiveness of downstream visual tasks. In such a context, it is particularly important to develop an algorithm that can automatically evaluate the image quality. According to whether the reference information of the original image is needed or not, objective image quality assessment methods can be categorized into full-reference IQA (FR-IQA), reduced-reference IQA (RR-IQA), and no-reference IQA (NR-IQA). In real scenarios, it is almost impossible to obtain the original reference image. Therefore, NR-IQA has become a popular topic in the field of computer vision and image processing.

Traditional NR-IQA methods map distorted images to objective quality scores by extracting hand-crafted features such as natural scene statistics (NSS) features\cite{mittal2012no,moorthy2011blind,saad2012blind}.  The limitation of the traditional method is that it does not perform well in practice when faced with complex and diverse distorted images. 
In recent years, deep learning-based image quality assessment methods\cite{wu2020end,pan2022dacnn,zhang2022dual,zhu2021generalizable,wang2022generation,gao2023blind} have attracted widespread attention due to their effectiveness. However, deep learning-based IQA methods are affected by the size of the dataset and data distribution. Current NR-IQA datasets are small in size and lack sufficient training samples. IQA models trained on small-scale datasets are inevitably biased, which limits their generalization and application in practical scenarios. Some established methods (e.g., CaHDC\cite{wu2020end}) expand the dataset by annotating the images using the FR-IQA model. However, the prediction results of the FR-IQA model may deviate from the subjective human perception scores, and there is still room for further improvement in the generalization ability of the trained model.

To further address the above problems and to efficiently improve the generalization of IQA models, many recent studies have explored multi-task strategy\cite{ma2017end,lin2018hallucinated,ren2018ran4iqa,yang2019sgdnet,li2021mmmnet,pan2022vcrnet}. This strategy aims to enhance the IQA task by integrating multiple networks and utilizing feature information from other tasks or domains. In this way, the model will be able to obtain richer feature representations from other networks and understand the features of the image more comprehensively. 
\IEEEpubidadjcol
The MEON\cite{ma2017end} method uses the distortion type identification subtask to extract features related to image distortion, thus providing effective crucial information for the quality assessment task. However, the method classifies specific distortion types from a set of predefined categories, limiting its generalization ability, especially for authentic distortion type images in wild scenes. To overcome this problem, we use a contrastive learning based approach to pre-train the distortion recognition sub-network. This approach enhances the network's ability to generalise to unknown distortion types, including authentic distortions. Li et al.\cite{li2021mmmnet} proposed MMMNet, which jointly optimises the IQA subtask and the saliency subtask, thus improving the saliency-guided IQA performance. 
Pan et al.\cite{pan2022vcrnet} proposed VCRNet, which is based on the task of visual restoration network. The performance of IQA is improved by combining the generated multilevel restoration features with the multilevel features generated by the quality assessment network. However, the limitation of these methods is that they usually perform multi-task feature fusion through simple operations (e.g., addition or concatenation) and provide only simple linear aggregation operations. To solve this problem, we adopt a feature fusion method based on the attention mechanism and propose a feature fusion module. The method effectively combines global contextual features with local contextual features and assigns appropriate weights to different features to better capture various details and distortions in the image.

In addition, the human visual system (HVS) has different perceptual sensitivities to information of different spatial frequencies in an image. Since the high frequency information reflects the texture and details of the image, HVS pays more attention to the high frequency content of the image. Existing IQA methods do not explicitly model the high frequency information of an image. Thus, we construct a specialized high-frequency extraction network to guide the IQA network to pay attention to the texture detail information of the image. 
In summary, the main contributions of this paper are as follows:
\begin{itemize}
\item[$\bullet$] We introduce a new multi-task based NR-IQA framework, consisting of a quality estimation, high-frequency extraction, and distortion-aware network for enhanced quality assessment.
\item[$\bullet$] We enhance the network's texture and detail perception by incorporating HVS-based high-frequency feature extraction, guiding the quality assessment to focus on important image details.
\item[$\bullet$] In order to better share and integrate information between multiple task networks and improve the network's ability to recognize distorted features, we adopt a feature fusion method based on the attention mechanism and propose a feature fusion module.
\end{itemize}

The remainder of this paper is organized as follows. Section II describes the related work on the NR-IQA approach. Section III describes our proposed end-to-end multi-task NR-IQA approach. Section IV gives the experimental results with related analysis. The paper is summarized in Section V.

\begin{figure*}[htbp]
\centerline{\includegraphics[width=16.5cm]{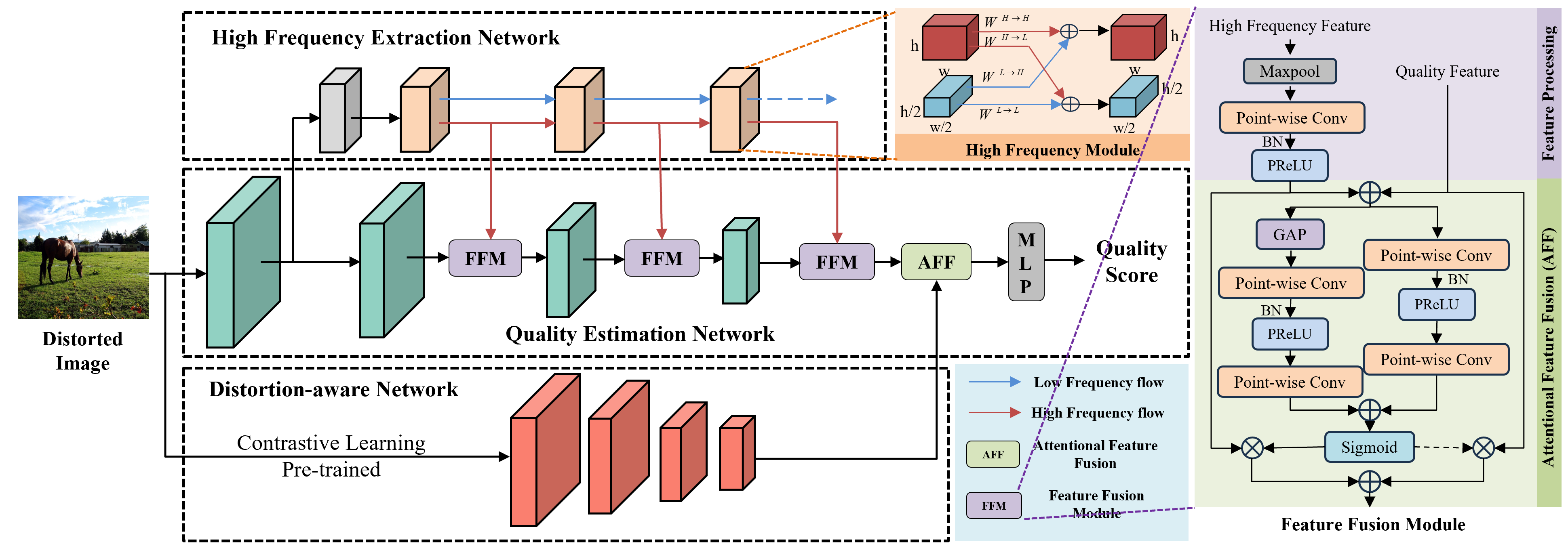}}
\caption{The framework of the proposed method. Our proposed framework consists of three branches: the Quality Estimation Network (QEN), the Distortion Aware Network (DAN), and the High Frequency Extraction Network (HFEN). The HFEN consists of vanilla convolution and high-frequency modules, with VAN as the backbone of QEN and ResNet-50 as the backbone of DAN. The QEN is the primary task, while the HFEN and the DAN are the auxiliary tasks. The DAN is pre-trained using Contrastive Learning to enhance the robustness of the distortion feature representation. The HFEN utilizes the high-frequency module to extract the high-frequency details, which helps the QEN to focus on key image components. In addition, an attention mechanism-based feature fusion method (AFF) is integrated for fusing distortion-aware features, and a feature fusion module (FFM) is proposed for adaptively fusing high-frequency features. }
\label{fig:frame}
\end{figure*}



\section{Related Works}
\subsection{NR-IQA based on traditional hand-crafted features}
Current generalized NR-IQA methods can be broadly classified into two categories: hand-crafted based methods and learning based methods. In traditional IQA methods, hand-crafted based approaches extract features manually through elaborate methods, such as natural scene statistics features. In early NSS-based methods, features are usually extracted through multiple transform domains, such as wavelet transform (WT) domain and discrete cosine transform (DCT) domain. Moorthy et al.\cite{moorthy2011blind} proposed DIIVINE, a wavelet transform-based method that first identifies the distortion type of a distorted image and then performs quality prediction for a specific distortion type. Saad et al.\cite{saad2012blind} proposed BLIINDS-II to develop an efficient generalized no-reference image quality assessment algorithm using a natural scene statistics (NSS) model with DCT coefficients. The method divides the image into equal sized localized image blocks and then performs DCT transform on each block. However, these methods result in high computational complexity and large computational time cost due to the domain transformations involved. Mittal et al.\cite{mittal2012no} proposed BRISQUE, a spatial domain-based method employing locally normalized luminance coefficients to quantify the natural loss by fitting the MSCN coefficient distributions of the image using an asymmetric generalized Gaussian distribution model. CORNIA \cite{ye2012unsupervised} uses a codebook-based approach to learn discriminative image features directly from raw image pixels, rather than using hand-crafted features.

\subsection{Deep learning based NR-IQA}
With the rise of deep learning, many deep learning methods are applied to IQA tasks. kang et al.\cite{kang2014convolutional} proposed an end-to-end CNN-based image quality assessment method. Thanks to the strong feature extraction capability of convolutional neural network, the method has gained great success compared to the traditional NR-IQA method. Liu et al.\cite{liu2017rankiqa} proposed a Siamese network based on Rank Learning to alleviate the problem of limited training samples for IQA. Zeng et al.\cite{zeng2017probabilistic} proposed a probabilistic quality representation (PQR) to approximate the distribution of subjective scores for each image, utilizing not only the MOS but also the standard deviation of the mean scores of the images. This enables the training of deep NR-IQA models that converge faster and are more stable even with a limited number of training samples. Bosse et al.\cite{bosse2017deep} proposed WaDIQaM, which predicts the score of each patch with corresponding weights by dividing the image into multiple patches. The method solves the problem that the quality of image patches is not uniformly distributed. Su et al.\cite{su2020blindly} proposed hyperIQA, which fuses local distortion features with global semantic features to aggregate fine-grained details and global information, enabling adaptive image quality assessment. 

\subsection{NR-IQA based on multi-task network}
In order to alleviate the problem of lack of sufficient labeled samples in the IQA database and to further improve the performance of NR-IQA, many recent studies have explored the strategy of multi-task networks. This strategy aims to enhance the quality assessment task by integrating multiple network streams and fully utilizing feature information from other tasks or domains. Ma et al.\cite{ma2017end} proposed MEON, which contains two subtasks, Subtask I categorizes an image into a specific type of distortion from a set of predefined categories, and Subtask II uses the distortion information obtained from Subtask I to predict the perceived quality of the same image. Lin et al.\cite{lin2018hallucinated} proposed Hallucinated-IQA, which includes a hallucinatory reference image generation subtask and a quality prediction subtask. The hallucinated reference generation subtask compensates for missing reference information using generative adversarial networks (GANs) to restore distorted images into hallucinated reference images for quality assessment. Yan et al.\cite{yan2018two} proposed a two-stream scheme to capture different levels of input information and to alleviate the difficulty of extracting features from one stream, with the gradient stream focusing on extracting detailed structural features and the image stream focusing on extracting intensity information. The method considers the inhomogeneity of the local distortion distribution in the image. li et al.\cite{li2021mmmnet} proposed MMMNet, an end-to-end multi-task deep convolutional neural network with multi-scale and multi-level fusion. Due to the visual attention of the human visual system, humans are more sensitive to aberrations in the attentional region than in the non-attentional focal region. The network jointly optimises the IQA subtask and saliency subtask to improve the performance of saliency-guided IQA. Pan et al.\cite{pan2022vcrnet} proposed VCRNet, which uses a nonadversarial model for the task of restoring distorted images, utilizing multiscale information from the image restoration process to aid in quality prediction. In 2023, Pan et al.\cite{pan2022no} proposed a multi-branch based NR-IQA method MBCNN. The method consists of two branches: spatial domain feature extraction and gradient domain feature extraction. The spatial domain feature extractor extracts distorted features. The gradient domain feature extractor is used to guide the spatial domain feature extractor to pay more attention to the distortion of structural information.

Although some NR-IQA methods based on multi-task networks have achieved bright results, there is still room for improvement. To enhance the model's effectiveness in concentrating on texture and detail information within the image, a high-frequency extraction network has been proposed to supplement the quality assessment network. The high-frequency extraction network is based on the priori knowledge of Human Visual System's (HVS) sensitivity to different image frequencies. With the objective of rendering high-frequency features more conducive to the feature fusion process, an alternative strategy has been implemented, setting it apart from standard practices like addition or concatenation. Concurrently, there has been the introduction of a Feature Fusion Module (FFM), which is underpinned by a feature fusion technique that operates on the principles of an attention mechanism. Different weights are assigned to the fused features by combining features from global context and local context to improve the representation of distortion information in IQA networks. In order to make the distortion perception task more effective in extracting the distortion information from the image, pre-training is performed using a contrastive learning method.



\section{Proposed Method}

Currently, one of the main problems of NR-IQA is that the database for IQA is small and lacks sufficient training samples. Models trained on small datasets are inevitably biased, limiting the application of the models. To solve this problem, some studies have proposed multi-task strategies, such as \cite{ma2017end}. With a multi-task strategy, features from other tasks or domains can be leveraged to enhance the quality assessment task. This approach helps alleviate the problem of insufficient labeled samples in the IQA database, as the model can learn more general features from large-scale data in other domains, rather than relying only on labeled image samples. In this paper, we present a novel end-to-end multi-task IQA solution. As shown in Fig.\ref{fig:frame}, our proposed method consists of three branches: a quality estimation network, a high-frequency extraction network, and a distortion-aware network. In this case, the quality estimation network serves as the main task, and the high frequency extraction network and the distortion perception network serve as auxiliary tasks. By leveraging feature information from other tasks or domains, the model is able to acquire richer feature representations. The combined use of main and auxiliary tasks can lead to more accurate and reliable image quality assessment.

In the proposed work process, given a distorted image $I \in \mathbb{R}^{3 \times w \times h}$, where $w$ and $h$ denote the width and height, and our goal is to predict its perceptual quality score. In the architecture of the quality estimation network, utilization has been made of a recently introduced visual attention network (VAN) \cite{guo2022visual}, intended for application in the domain of visual tasks. This VAN is positioned as the backbone of the network. The attention mechanism within this network operates through decomposed large kernel convolution, an approach that adeptly secures the representation of long-term dependencies inherent in image features, concurrently achieving a decrease in the computational requirements. 
Image $I$ is first fed into the quality estimation network. Let $f_{i}$ denotes the $i$th block of the backbone network, $g_{i}$ denotes the feature output from the $i$th high frequency module of the high frequency extraction network. After the first block of backbone network, the feature $F_{1}$ is obtained, then $F_{1}$ is fed into the high-frequency extraction network. The whole process can be described as follows:
\begin{eqnarray}
F_{i+1} & = & f_{i+1}\left(F_{i}\right) \oplus g_{i}\left(f_{1}(I)\right)
\end{eqnarray}
where $i \in\{1,2,3\}$, $F_{i+1}$ denotes the output fused features and $\oplus$ denotes the feature fusion module. In order to effectively fuse features from different task network streams, we adopt an attention-based feature fusion method AFF\cite{dai2021attentional}, based on which we propose a feature fusion module (FFM) to fuse high-frequency features and multilevel quality features in quality estimation network. Finally, quality features fused with high-frequency information are fused with features from the distortion-aware network and fed to the MLP head to predict perceptual quality scores.

\subsection{High Frequency Extraction Network}
The human visual system (HVS) has different perceptual sensitivities to information of different spatial frequencies in images\cite{de1980spatial}. In the field of image quality assessment, taking into account the differences in the perception of low and high frequency information by the HVS can help design algorithms that better simulate human sensitivity to image details. High-frequency information usually reflects the texture and details of an image, while HVS tends to focus more on the high-frequency part of an image when observing it. 
According to the spatial frequency model of vision \cite{campbell1968application}, a natural image can be decomposed into a low-frequency part and a high-frequency part. The low-frequency part usually reflects areas of the image that change slowly, and the high-frequency part usually reflects details or edges in the image. The design of a method for the decomposition of high-frequency components has been executed, with the utilization of octave convolution \cite{chen2019drop} facilitating the segregation of distorted image features into two distinct categories: those of a high-frequency nature and those of a low-frequency nature. The high-frequency features are utilized to guide the quality estimation network for quality assessment. The proposed high-frequency extraction network, which comprises Vanilla convolution and high-frequency extraction modules, is designed to extract valuable high-frequency information. This design also serves to reduce computational complexity, as the low-frequency part is discarded. The flow of the high frequency extraction module is shown below:
\begin{eqnarray}
\begin{aligned}
F_{i+1}^{H}& = &f\left(F_{i}^{H} ; W_{i}^{H \rightarrow H}\right)+\text {up}\left(f\left(F_{i}^{L} ; W_{i}^{L \rightarrow H}\right), 2\right) \\
F_{i+1}^{L}& = &f\left(F_{i}^{L} ; W_{i}^{L \rightarrow L}\right)+f\left(\text {down}\left(F_{i}^{H}, 2\right) ; W_{i}^{H \rightarrow L}\right)
\end{aligned}
\end{eqnarray}
where $F_{i+1}^{L}$ and $F_{i+1}^{H}$ denote the low-frequency and high-frequency features, respectively, and $f(F, W)$ denotes Convolution operation with parameter $W$. $up(F,2)$ denotes upsampling using nearest interpolation method, and $down(F,2)$ denotes twice the downsampling using the average pooling layer.

\subsection{Feature Fusion Module (FFM)}
Previous feature fusion methods for IQA are usually realized by simple operations. The notion persists that straightforward procedures like addition or concatenation are confined to providing basic linear aggregation operations. Such operations are considered to be potentially detrimental to the expressive power required for capturing feature distortions, thereby rendering them less suitable for applications within the domain of image quality assessment. To achieve a more effective fusion of high-frequency features and a more expressive representation of image features, a feature fusion approach has been embraced, drawing inspiration from the work detailed in \cite{dai2021attentional}. This approach is founded upon the principles of the attention mechanism. Concurrently, there has been an introduction of a novel component known as the Feature Fusion Module (FFM). 

Fig.\ref{fig:frame} depicts the structure of the feature fusion module, which includes feature processing and attentional feature fusion (AFF). For the high-frequency features derived from the high-frequency extraction network, it is believed that downsampling via the maxpool layer can effectively encapsulate the high-frequency information of the image. So first the features are downsampled to the same size as the quality features using the maxpool layer, followed by processing the channel dimensions using point-wise convolution of 1×1 size. Subsequently, batch normalization \cite{ioffe2015batch} and PReLU layer \cite{he2015delving} are used to increase the nonlinear relationship between the layers and accelerate the training process. Subsequently, the processed high-frequency features and the quality features, which are derived from the quality estimation network, are input into AFF, a feature fusion module that operates on the basis of the attention mechanism. In AFF, the two parts of the features are first summed up and then fed into two branches to extract the local contextual information and global contextual information of the features, respectively. After the local and global features are aggregated, the weights of the features are calculated by sigmoid function. The channel attention at both local and global scales is realized. The whole process can be described as follows:
\begin{eqnarray}
\begin{aligned}
F_{h f}=f_{P R e L U}\left(f_{B N}\left(f_{\text {Conv }}\left(f_{\text {down }}\left(F_{h f}^{0}\right)\right)\right)\right)  \\
F_{\text {output }}=w\left(F_{h f} \oplus F_{q f}\right) \otimes F_{h f} \\
+(1-w)\left(F_{h f} \oplus F_{q f}\right) \otimes F_{q f}
\end{aligned}
\end{eqnarray}
where $F_{h f}^{0}$ denotes the high frequency features and $F_{q f}$ denotes the quality features. $f_{\text {down }}$ denotes the max pooling operation, $f_{\text {Conv }}$ denotes the point-wise convolution, $f_{B N}$ denotes the batch normalization operation, and $f_{P R e L U}$ denotes the PReLU activation function, $\oplus$ denotes broadcast addition, $\otimes$ denotes element-wise multiplication. $w$ denotes the feature weights computed by the sigmoid function in the AFF, and the computation process of $w$ can be expressed as follows:

\begin{eqnarray}
\begin{aligned}
w = f_{\text {sigmoid }}\left(f_{\text {Conv }}\left(f_{B N}\left(f_{\text {Conv }}\left(f_{G A P}\left(F_{\text {add }}\right)\right)\right)\right) \oplus\right. \\
\left.f_{\text {Conv }}\left(f_{\text {BN }}\left(f_{\text {Conv }}\left(F_{\text {add }}\right)\right)\right)\right)
\end{aligned}
\end{eqnarray}
where $f_{\text {sigmoid }}$ denotes the sigmoid activation function and $F_{add}$ denotes $F_{h f} \oplus F_{q f}$.

\subsection{Distortion-Aware Network}
Established methods \cite{ma2017end} train a distortion type recognition sub-network by a supervised approach, which then aids in quality assessment. This approach categorizes specific distortion types from a set of predefined categories. It is limited in that it can only be effective for a limited number of synthetic distortion types, and its performance is greatly reduced for authentic distortion types in the wild images. The currently available annotated IQA datasets have a limited number of samples, and annotating the data is expensive and time-consuming, which leads us to consider unsupervised training approaches. Owing to the broad utilization of contrastive learning methodologies within the realm of computer vision, the extraction of image distortion information has been facilitated through the application of a contrastive learning method. The backbone network for the distortion-aware subtask is constituted by the ResNet-50 encoder that has been trained through the processes outlined in CONTRIQUE\cite{madhusudana2022image}.
For synthetic distorted images, the method employs known distortion types as classification labels and establishes associations by measuring the two-by-two similarity between each pair of images in the batch. Specifically, distorted images with the same distortion type and the same degradation level are classified into the same category, resulting in a pre-training dataset for contrastive learning. This approach is intended to enable the model to learn to capture the relationship between similar distortion types and similar quality grade levels, providing effective information for subsequent image quality assessment tasks. The synthetic distortion loss function is the normalized temperature-scale cross-entropy (NT-Xent) loss, which is defined for the input image $x_i$ as:
\begin{eqnarray}
\begin{aligned}
\mathcal{L}_{i}^{s y n} = \frac{1}{|P(i)|} \sum_{j \in P(i)}-\log \frac{\exp \left(\phi\left(z_{i}, z_{j}\right) / \tau\right)}{\sum_{k = 1}^{N} \|_{k \neq i} \exp \left(\phi\left(z_{i}, z_{k}\right) / \tau\right)}
\end{aligned}
\end{eqnarray}
where $N$ denotes the batch size of the image, $z$ denotes the feature vector of the encoder output, $\phi(\cdot)$ denotes the vector dot product operation, $\|$ denotes the indicator function, $\tau$ denotes the temperature parameter, and $P(i)$ refers to the set of images belonging to the same class as $x_i$. For authentic distorted images in the wild, each authentic distorted image is classified into a separate category because distortion in wild images is usually a complex interaction between various distortions. The loss function for authentic distorted images is defined as:
\begin{eqnarray}
\begin{aligned}
\mathcal{L}_{i}^{\text {aut }}=-\log \frac{\exp \left(\phi\left(z_{i}, z_{j}\right) / \tau\right)}{\sum_{k=1}^{N} \|_{k \neq i} \exp \left(\phi\left(z_{i}, z_{k}\right) / \tau\right)}
\end{aligned}
\end{eqnarray}
In the image data enhancement stage, each image has two transform versions, full scale, and half scale transforms, resulting in data samples where at least two distorted images belong to the same class. By introducing the half-scale transformation, the richness of the dataset can be increased while maintaining the integrity of the image information, thus improving the generalization ability and robustness of the model. The total pretraining loss is defined as:
\begin{eqnarray}
\begin{aligned}
\mathcal{L}=\frac{1}{N} \sum_{i=1}^{N}\left\|_{\left(x_{i} \text { AAuhhentic }\right)} \mathcal{L}_{i}^{\text {syn }}+\right\|_{\left(x_{i} \in \text { Authentic }\right)} \mathcal{L}_{i}^{\text {aut }}
\end{aligned}
\end{eqnarray}
where $\|$ is an indicator function to determine whether the input image is true distortion or not. The encoder trained by this contrastive learning strategy can understand the relationship between different distorted images more comprehensively and improve the generalization ability of the image quality assessment model. It should be noted that the approach diverges from the one outlined in CONTRIQUE, wherein the original-sized image is opted for as the direct input, bypassing the use of features derived from multiple scales. The utilization of features extracted from an image at its original dimensions following the processing by a pre-trained encoder is considered to result in a comprehensive depiction of the distorted features.

\section{Experiments and Analyses}

\begin{table*}
\centering
\caption{Detailed information of five common datasets}
\label{table:1}
\begin{tabular}{c|cccccc}
\hline
Dataset   & Reference Images & Distorted Images & Distortion Type & Distortion Types & Score Type & Score Range \\ \hline
LIVE\cite{sheikh2006statistical}      & 29               & 779              & synthetic       & 5                & DMOS       & {[}0,100{]} \\
CSIQ\cite{larson2010most}      & 30               & 866              & synthetic       & 6                & DMOS       & {[}0,1{]}   \\
TID2013\cite{ponomarenko2015image}   & 25               & 3000             & synthetic       & 24               & MOS        & {[}0,9{]}   \\
CLIVE\cite{ghadiyaram2015massive}     & N/A              & 1162             & authentic       & N/A              & MOS        & {[}0,100{]} \\
KonIQ\cite{hosu2020koniq} & N/A              & 10073            & authentic       & N/A              & MOS        & {[}1,5{]}   \\ \hline
\end{tabular}
\end{table*}

\subsection{Experimental Setup}
To effectively evaluate the performance of our proposed model, five common IQA datasets were utilized. These include three synthetic distortion datasets (LIVE\cite{sheikh2006statistical}, CSIQ\cite{larson2010most}, TID2013\cite{ponomarenko2015image}) and two authentic distortion datasets (LIVEC\cite{ghadiyaram2015massive}, KONIQ\cite{hosu2020koniq}). Details of the five common datasets are shown in Table \ref{table:1}, where MOS denotes mean opinion score and DMOS denotes difference mean opinion score. 

\begin{table*}
\centering
\caption{Comparison of our proposed method with SOTA algorithms on three synthetically distortion datasets and two authentically distortion datasets.}
\label{table:2}
\begin{tabular}{c|cc|cc|cc|cc|cc}
\hline
                   & \multicolumn{2}{c|}{LIVE}                              & \multicolumn{2}{c|}{CSIQ}                                                     & \multicolumn{2}{c|}{TID2013}                                                  & \multicolumn{2}{c|}{LIVEC}                             & \multicolumn{2}{c}{KonIQ}                              \\ \cline{2-11} 
\multirow{-2}{*}{} & PLCC           & SRCC                                  & PLCC                                  & SRCC                                  & PLCC                                  & SRCC                                  & PLCC                                  & SRCC           & PLCC           & SRCC                                  \\ \hline
DIIVINE\cite{moorthy2011blind}            & 0.908          & 0.892                                 & 0.776                                 & 0.804                                 & 0.567                                 & 0.643                                 & 0.591                                 & 0.588          & 0.558          & 0.546                                 \\
BRISQUE\cite{mittal2012no}            & 0.944          & 0.929                                 & 0.748                                 & 0.812                                 & 0.571                                 & 0.626                                 & 0.629                                 & 0.629          & 0.685          & 0.681                                 \\
CORNIA\cite{ye2012unsupervised}             & 0.938          & 0.937                                 & 0.768                                 & 0.695                                 & 0.734                                 & 0.657                                 & 0.636                                 & 0.617          & 0.569          & 0.545                                 \\
HOSA\cite{xu2016blind}               & 0.943          & 0.942                                 & 0.805                                 & 0.721                                 & 0.835                                 & 0.749                                 & 0.657                                 & 0.621          & 0.703          & 0.684                                 \\
MEON\cite{ma2017end}               & 0.955          & 0.951                                 & 0.864                                 & 0.852                                 & 0.824                                 & 0.808                                 & 0.710                                 & 0.697          & 0.628          & 0.611                                 \\
WaDIQaM\cite{bosse2017deep}            & 0.955          & 0.960                                 & 0.844                                 & 0.852                                 & 0.855                                 & 0.835                                 & 0.671                                 & 0.682          & 0.807          & 0.804                                 \\
TS-CNN\cite{yan2018two}             & 0.965          & 0.969                                 & 0.904                                 & 0.892                                 & 0.824                                 & 0.783                                 & 0.687                                 & 0.654          & 0.724          & 0.713                                 \\
RAN4IQA\cite{ren2018ran4iqa}            & 0.962          & 0.961                                 & 0.931                                 & 0.914                                 & 0.859                                 & 0.820                                 & 0.612                                 & 0.586          & 0.763          & 0.752                                 \\
TRIQ\cite{you2021transformer}               & 0.965          & 0.949                                 & 0.838                                 & 0.825                                 & 0.858                                 & 0.846                                 & 0.861                                 & 0.845          & 0.903          & 0.892                                 \\
MetaIQA\cite{zhu2020metaiqa}            & 0.959          & 0.960                                 & 0.908                                 & 0.899                                 & 0.868                                 & 0.856                                 & 0.802                                 & 0.835          & 0.856          & 0.887                                 \\
HyperIQA\cite{su2020blindly}           & 0.966          & 0.962                                 & 0.942                                 & 0.923                                 & 0.858                                 & 0.840                                 & \textbf{0.882}                        & 0.859          & 0.917          & 0.906                                 \\
CaHDC\cite{wu2020end}              & 0.964          & 0.965                                 & 0.914                                 & 0.903                                 & 0.885                                 & 0.862                                 & 0.744                                 & 0.738          & /              & /                                     \\
TReS\cite{golestaneh2022no}               & 0.968          & 0.969                                 & 0.942                                 & 0.922                                 & 0.883                                 & 0.863                                 & 0.877                                 & 0.846          & \textbf{0.928} & {\color[HTML]{000000} \textbf{0.915}} \\
MMMNet\cite{li2021mmmnet}             & 0.970          & 0.970                                 & {\color[HTML]{000000} \textbf{0.937}} & {\color[HTML]{000000} \textbf{0.924}} & 0.853                                 & 0.832                                 & 0.876                                 & 0.852          & /              & /                                     \\
VCRNet\cite{pan2022vcrnet}             & \textbf{0.973} & \textbf{0.975}                        & 0.932                                 & 0.914                                 & {\color[HTML]{000000} \textbf{0.886}} & {\color[HTML]{000000} \textbf{0.865}} & 0.878                                 & \textbf{0.864} & 0.919          & 0.906                                 \\
\textbf{proposed}  & \textbf{0.973} & {\color[HTML]{000000} \textbf{0.974}} & \textbf{0.967}                        & \textbf{0.962}                        & \textbf{0.916}                        & \textbf{0.897}                        & {\color[HTML]{000000} \textbf{0.880}} & \textbf{0.864} & \textbf{0.928} & \textbf{0.919}                        \\ \hline
\end{tabular}
\end{table*}

The experiments are trained and tested on NVIDIA GeForce RTX 4090, and our proposed method is implemented through the PyTorch framework. For data preprocessing, patches from each image are randomly selected to cover a significant portion of the original distorted image, and these patches are subjected to random horizontal and vertical flipping for the purpose of data augmentation. Aggressive data augmentation strategies are avoided, and it should be noted that the flipping strategy used does not affect the quality of the images. The standard training strategy\cite{golestaneh2022no} for existing IQA algorithms was employed. 
During the procedure, a total of fifty patches, with each patch being a square of $224 \times 224$ pixels in dimension, are randomly chosen from every image that is part of the training dataset.
The quality scores from the original distorted images are assumed to be inherited by these patches. During training, the learning rate of the model was set to an initial value of $2e-5$ and reduced by a factor of $10$ for each subsequent epoch. The Adam optimizer was used with weight decay of $5e-4$ and batch size set to $64$. During the model's testing phase, $224 \times 224$ patches are cropped randomly from each test image. Subsequently, the quality scores that are forecasted by each individual patch undergo a process of averaging, culminating in the determination of the ultimate quality score for the image. The common IQA experimental approach was followed\cite{golestaneh2022no}, and the same setup was adopted for all experiments. In the dataset, $80\%$ of the distorted images were used in the training phase, while the remaining $20\%$ were set aside for the testing phase. 
To eliminate the effect of sampling bias in the dataset, each database was randomly split $10$ times for all experiments. 


For the assessment of model performance, two commonly used metrics were employed: the Spearman rank-order correlation coefficient (SRCC) \cite{puth2015effective} and the Pearson linear correlation coefficient (PLCC) \cite{sedgwick2012pearson}. SRCC reflects the relationship between objective quality scores and predicted scores, and PLCC reflects the linear relationship between objective quality scores and predicted scores. The formula for SRCC is:
\begin{align}
S R C C = 1-\frac{6 \sum_{i = 1}^{B}\left(x_{i}-y_{i}\right)^{2}}{B\left(B^{2}-1\right)}
\end{align}
where $B$ denotes the number of samples, and $x_{i}$ and $y_{i}$ denote the subjective quality score ranking and objective score ranking of image $i$, respectively. The formula for PLCC is:
\begin{align}
P L C C & = \frac{\sum_{i}\left(s_{i}-\hat{s}\right)\left(o_{i}-\hat{o}\right)}{\sqrt{\sum_{i}\left(s_{i}-\hat{s}\right)^{2}} \sqrt{\sum_{i}\left(o_{i}-\hat{o}\right)^{2}}}
\end{align}
where $s_{i}$ and $o_{i}$ denote the predicted and objective quality scores, respectively, and $\hat{s}$ and $\hat{o}$ denote the corresponding mean scores, respectively.
\begin{table*}
\centering
\caption{Performance of single distortion types in the LIVE dataset.}
\label{table:3}
\begin{tabular}{c|ccccc|ccccc}
\hline
                         & \multicolumn{5}{c|}{SRCC}                                                                                                                               & \multicolumn{5}{c}{PLCC}                                                                                                                                                                              \\ \cline{2-11} 
\multirow{-2}{*}{Method} & JP2K                                  & JPEG           & WN             & GBLUR                                 & FF                                    & JP2K                                  & JPEG                                  & WN                                    & GBLUR                                 & FF                                    \\ \hline
DIIVINE\cite{moorthy2011blind}                  & 0.925                                 & 0.913          & \textbf{0.985} & 0.789                                 & 0.873                                 & 0.901                                 & 0.887                                 & {\color[HTML]{000000} \textbf{0.987}} & 0.787                                 & 0.879                                 \\
BRISQUE\cite{mittal2012no}                  & 0.914                                 & 0.965          & 0.979          & 0.951                                 & 0.877                                 & 0.923                                 & 0.973                                 & 0.985                                 & 0.951                                 & 0.903                                 \\
CORNIA\cite{ye2012unsupervised}                   & 0.936                                 & 0.934          & 0.962          & 0.926                                 & 0.912                                 & 0.924                                 & 0.906                                 & 0.945                                 & 0.934                                 & 0.913                                 \\
HOSA\cite{xu2016blind}                     & 0.928                                 & 0.936          & 0.964          & 0.964                                 & 0.934                                 & 0.923                                 & 0.924                                 & 0.959                                 & 0.965                                 & 0.923                                 \\
CNN\cite{kang2014convolutional}                      & 0.936                                 & 0.965          & 0.974          & 0.952                                 & 0.906                                 & 0.942                                 & 0.973                                 & 0.976                                 & 0.953                                 & 0.907                                 \\
MEON\cite{ma2017end}                     & 0.965                                 & 0.954          & 0.974          & 0.941                                 & 0.885                                 & 0.965                                 & 0.955                                 & 0.978                                 & 0.942                                 & 0.868                                 \\
WaDIQaM\cite{bosse2017deep}                  & 0.962                                 & 0.956          & 0.969          & 0.946                                 & 0.927                                 & 0.965                                 & 0.960                                 & 0.973                                 & 0.957                                 & 0.936                                 \\
TS-CNN\cite{yan2018two}                   & 0.959                                 & 0.965          & 0.981          & \textbf{0.970}                        & 0.930                                 & 0.962                                 & 0.970                                 & \textbf{0.990}                        & \textbf{0.973}                        & 0.943                                 \\
CaHDC\cite{wu2020end}                    & 0.948                                 & 0.970          & 0.978          & 0.951                                 & 0.898                                 & 0.953                                 & 0.973                                 & 0.982                                 & 0.955                                 & 0.913                                 \\
HyperIQA\cite{su2020blindly}                 & {\color[HTML]{000000} \textbf{0.965}} & 0.956          & 0.974          & 0.963                                 & 0.923                                 & \textbf{0.973}                        & 0.962                                 & 0.975                                 & {\color[HTML]{000000} \textbf{0.971}} & 0.935                                 \\
MMMNet\cite{li2021mmmnet}                   & \textbf{0.968}                        & \textbf{0.974} & \textbf{0.985} & 0.935                                 & {\color[HTML]{000000} \textbf{0.936}} & /                                     & /                                     & /                                     & /                                     & /                                     \\
VCRNet\cite{pan2022vcrnet}                   & 0.964                                 & 0.970          & 0.972          & 0.955                                 & 0.935                                 & {\color[HTML]{000000} \textbf{0.972}} & {\color[HTML]{000000} \textbf{0.984}} & 0.981                                 & 0.968                                 & \textbf{0.951}                        \\
proposed                 & 0.957                                 & \textbf{0.974} & 0.980          & {\color[HTML]{000000} \textbf{0.965}} & \textbf{0.947}                        & 0.970                                 & \textbf{0.988}                        & {\color[HTML]{000000} \textbf{0.987}} & 0.960                                 & {\color[HTML]{000000} \textbf{0.945}} \\ \hline
\end{tabular}
\end{table*}
\subsection{Performance Comparison with SOTA Methods}
To validate the superior performance of our proposed method, we conducted a comparative analysis using a curated set of $15$ quintessential IQA models, including $4$ hand-crafted-based methods, DIIVINE, BRISQUE, CORNIA, and HOSA, and 11 deep-learning-based methods, MEON, WaDIQaM, TS-CNN, RAN4IQA, TIQA, MetaIQA, HyperIQA, CaHDC, TReS, MMMNet, and VCRNet. The experimental results are shown in the table \ref{table:2}, with the first two results in bold. The results presented in Table \ref{table:2} allow for the following conclusions to be derived: First and foremost, it is determined that the proposed method occupies the optimal tier within the spectrum of the majority of methods. Out of a total of $10$ metrics in the five datasets, the proposed method achieved $8$ optimal results and $2$ competitive results. Specifically, the proposed model achieves $3.0\%$, $3.8\%$ (PLCC, SRCC) higher than MMMNet(second-best) in CSIQ dataset and $3.0\%$, $3.2\%$ (PLCC, SRCC)  higher than VCRNet in TID2013 dataset. Second, in the small dataset LIVE, the optimal level was not reached by the proposed method. This result can be attributed to the heightened overall network complexity inherent in the multi-task approach. Despite this, the method still managed to achieve results that are considered competitive. The performance of the proposed method on the LIVE dataset differs from VCRNet(best) by only $0.1\%$ (SRCC). The superiority of the proposed method is demonstrated by its excellent performance on larger scale datasets (TID2013,KONIQ). Finally, at the dataset level, although the proposed method is not optimized for specific synthetic and authentic data, the proposed method still performs well on synthetic datasets (CSIQ, TID2013) versus authentic datasets (LIVEC, KONIQ). Specifically, our proposed method achieves the best results of $0.916$, $0.897$ (PLCC,SRCC) in TID2013 and $0.928$, $0.919$ (PLCC,SRCC) in KONIQ.

\begin{table*}
\centering
\caption{Performance of single distortion types in the CSIQ dataset.}
\label{table:4}
\begin{tabular}{c|cccccc|cccccc}
\hline
                         & \multicolumn{6}{c|}{SRCC}                                                                                                                                                                                                                     & \multicolumn{6}{c}{PLCC}                                                                                                                                                                                                                      \\ \cline{2-13} 
\multirow{-2}{*}{Method} & WN                                    & JPEG                                  & JP2K                                  & PN                                    & BLUR                                  & CC                                    & WN                                    & JPEG                                  & JP2K                                  & PN                                    & BLUR                                  & CC                                    \\ \hline
DIIVINE\cite{moorthy2011blind}                  & 0.756                                 & 0.732                                 & 0.803                                 & 0.432                                 & 0.785                                 & 0.789                                 & 0.783                                 & 0.758                                 & 0.844                                 & 0.613                                 & 0.836                                 & 0.806                                 \\
BRISQUE\cite{mittal2012no}                  & 0.723                                 & 0.806                                 & 0.840                                 & 0.378                                 & 0.820                                 & 0.804                                 & 0.742                                 & 0.828                                 & 0.887                                 & 0.496                                 & 0.891                                 & 0.835                                 \\
CORNIA\cite{ye2012unsupervised}                   & 0.664                                 & 0.513                                 & 0.831                                 & 0.493                                 & 0.836                                 & 0.462                                 & 0.687                                 & 0.563                                 & 0.883                                 & 0.632                                 & 0.904                                 & 0.543                                 \\
HOSA\cite{xu2016blind}                     & 0.604                                 & 0.733                                 & 0.818                                 & 0.500                                 & 0.841                                 & 0.716                                 & 0.656                                 & 0.759                                 & 0.899                                 & 0.601                                 & 0.912                                 & 0.744                                 \\
CNN\cite{kang2014convolutional}                      & 0.919                                 & 0.915                                 & 0.930                                 & 0.900                                 & 0.918                                 & 0.786                                 & 0.945                                 & 0.944                                 & 0.945                                 & 0.834                                 & {\color[HTML]{000000} \textbf{0.939}} & 0.784                                 \\
MEON\cite{ma2017end}                     & 0.947                                 & {\color[HTML]{000000} \textbf{0.946}} & 0.896                                 & 0.895                                 & 0.908                                 & 0.807                                 & 0.952                                 & 0.967                                 & 0.921                                 & 0.895                                 & {\color[HTML]{000000} \textbf{0.939}} & 0.814                                 \\
WaDIQaM\cite{bosse2017deep}                  & 0.947                                 & 0.926                                 & 0.946                                 & 0.879                                 & 0.909                                 & 0.869                                 & {\color[HTML]{000000} \textbf{0.956}} & 0.934                                 & 0.957                                 & 0.886                                 & 0.916                                 & 0.873                                 \\
TS-CNN\cite{yan2018two}                   & {\color[HTML]{000000} \textbf{0.957}} & 0.940                                 & 0.941                                 & 0.931                                 & 0.906                                 & 0.818                                 & {\color[HTML]{000000} \textbf{0.956}} & 0.971                                 & 0.947                                 & 0.933                                 & 0.923                                 & 0.807                                 \\
CaHDC\cite{wu2020end}                    & 0.896                                 & 0.900                                 & 0.936                                 & 0.874                                 & 0.912                                 & 0.872                                 & 0.912                                 & 0.924                                 & 0.943                                 & 0.896                                 & 0.923                                 & 0.879                                 \\
HyperIQA\cite{su2020blindly}                 & 0.936                                 & 0.945                                 & {\color[HTML]{000000} \textbf{0.956}} & 0.935                                 & {\color[HTML]{000000} \textbf{0.922}} & 0.892                                 & 0.942                                 & 0.946                                 & {\color[HTML]{000000} \textbf{0.959}} & 0.946                                 & 0.924                                 & 0.897                                 \\
MMMNet\cite{li2021mmmnet}                   & 0.879                                 & 0.912                                 & 0.932                                 & 0.941                                 & 0.894                                 & \textbf{0.942}                        & /                                     & /                                     & /                                     & /                                     & /                                     & /                                     \\
VCRNet\cite{pan2022vcrnet}                   & 0.883                                 & 0.919                                 & 0.931                                 & {\color[HTML]{000000} \textbf{0.958}} & 0.912                                 & {\color[HTML]{000000} \textbf{0.938}} & 0.897                                 & {\color[HTML]{000000} \textbf{0.979}} & {\color[HTML]{000000} \textbf{0.959}} & {\color[HTML]{000000} \textbf{0.952}} & 0.925                                 & \textbf{0.935}                        \\
proposed                 & \textbf{0.972}                        & \textbf{0.967}                        & \textbf{0.968}                        & \textbf{0.961}                        & \textbf{0.960}                        & 0.877                                 & \textbf{0.973}                        & \textbf{0.993}                        & \textbf{0.967}                        & \textbf{0.966}                        & \textbf{0.957}                        & {\color[HTML]{000000} \textbf{0.924}} \\ \hline
\end{tabular}
\end{table*}
\subsection{Performance Evaluation of Individual Distortion Types}

To evaluate and compare performance across various image distortion types, we conducted additional experiments using the LIVE, CSIQ, and TID2013 datasets for comparative analysis. Across all utilized databases, the training process involves exposure to all types of distortions, whereas the testing phase is confined to evaluating against a single type of distortion. The training data is not duplicated with the test data. The methods compared include some traditional IQA methods (e.g., DIIVINE, BRISQUE) and recently deep learning-based methods (e.g., TS-CNN, MMMNet, and VCRNet). In Table \ref{table:3}, we evaluate the performance of LIVE dataset for all distortion types, which are Jpeg2000 compression, jpeg compression, Gaussian white noise, Gaussian blur and Fast Fading Rayleigh. From the results in the table \ref{table:3}, it can be observed that our proposed method achieves competitive results in most distortion types. Specifically, the proposed method performs well in Jpeg compression and Fast Fading Rayleigh, reaching $0.974$, $0.988$ (SRCC, PLCC) and $0.947$, $0.945$ (SRCC, PLCC), respectively. The effectiveness of this result is due to the fact that high-frequency information inherent in the two distortion types is efficiently harnessed by our proposed high-frequency extraction network. Concurrently, it should be noted that there is a noted underperformance of the proposed method when it comes to handling distortions of the Gaussian blur type. The potential reason for the model's suboptimal performance could be identified as the reduced presence of high-frequency information in blurred images, leading to an ineffective utilization of information by the model. 

\begin{table*}
\centering
\caption{Performance of single distortion types in the TID2013 dataset.}
\label{table:5}
\begin{tabular}{l|ccccccccc}
\hline
SRCC                                 & DIIVINE & BRISQUE & CORNIA & HOSA                                  & RAN4IQA                               & SGDNet                                & MMMNet                                & VCRNet                                & Proposed                              \\ \hline
additive gayssian noise              & 0.785   & 0.879   & 0.655  & 0.817                                 & 0.866                                 & 0.668                                 & {\color[HTML]{000000} \textbf{0.833}} & 0.800                                 & \textbf{0.874}                        \\
additive noise in color   components & 0.493   & 0.536   & 0.164  & 0.553                                 & {\color[HTML]{000000} \textbf{0.753}} & 0.579                                 & 0.614                                 & 0.749                                 & \textbf{0.784}                        \\
spatially correlated noise           & 0.833   & 0.782   & 0.709  & {\color[HTML]{000000} \textbf{0.881}} & 0.842                                 & 0.704                                 & 0.869                                 & 0.875                                 & \textbf{0.902}                        \\
masked noise                         & 0.406   & 0.260   & 0.463  & 0.556                                 & 0.462                                 & 0.473                                 & 0.541                                 & {\color[HTML]{000000} \textbf{0.627}} & \textbf{0.872}                        \\
high frequency noise                 & 0.836   & 0.905   & 0.851  & 0.865                                 & {\color[HTML]{000000} \textbf{0.908}} & 0.818                                 & 0.791                                 & 0.894                                 & \textbf{0.930}                        \\
impulse noise                        & 0.723   & 0.839   & 0.596  & 0.859                                 & 0.855                                 & 0.719                                 & 0.687                                 & {\color[HTML]{000000} \textbf{0.912}} & \textbf{0.963}                        \\
quantization noise                   & 0.652   & 0.681   & 0.703  & 0.679                                 & 0.849                                 & 0.644                                 & 0.767                                 & {\color[HTML]{000000} \textbf{0.901}} & \textbf{0.960}                        \\
gaussian blur                        & 0.833   & 0.842   & 0.866  & 0.870                                 & 0.833                                 & 0.892                                 & {\color[HTML]{000000} \textbf{0.926}} & 0.903                                 & \textbf{0.945}                        \\
image denoising                      & 0.711   & 0.565   & 0.736  & 0.845                                 & 0.839                                 & {\color[HTML]{000000} \textbf{0.874}} & 0.732                                 & \textbf{0.921}                        & 0.838                                 \\
jpeg compression                     & 0.818   & 0.831   & 0.805  & 0.896                                 & {\color[HTML]{000000} \textbf{0.939}} & 0.885                                 & 0.889                                 & 0.916                                 & \textbf{0.960}                        \\
jpeg2000 compression                 & 0.816   & 0.756   & 0.855  & 0.901                                 & \textbf{0.912}                        & 0.897                                 & 0.889                                 & {\color[HTML]{000000} \textbf{0.907}} & 0.891                                 \\
jpeg transmission errors             & 0.346   & 0.278   & 0.663  & 0.655                                 & 0.566                                 & 0.712                                 & 0.776                                 & \textbf{0.795}                        & {\color[HTML]{000000} \textbf{0.780}} \\
jpeg2000 compression errors          & 0.817   & 0.703   & 0.677  & 0.683                                 & 0.772                                 & 0.722                                 & {\color[HTML]{000000} \textbf{0.916}} & 0.848                                 & \textbf{0.946}                        \\
non-eccentricity pattern noise       & 0.174   & 0.264   & 0.199  & 0.176                                 & 0.234                                 & 0.283                                 & 0.630                                 & {\color[HTML]{000000} \textbf{0.668}} & \textbf{0.785}                        \\
local block-wise distortions         & 0.366   & 0.272   & 0.021  & 0.269                                 & 0.339                                 & 0.612                                 & 0.546                                 & \textbf{0.716}                        & {\color[HTML]{000000} \textbf{0.640}} \\
mean shift                           & 0.106   & 0.239   & 0.203  & 0.128                                 & 0.135                                 & 0.342                                 & 0.392                                 & {\color[HTML]{000000} \textbf{0.528}} & \textbf{0.544}                        \\
contrast change                      & 0.132   & 0.090   & 0.254  & 0.138                                 & 0.578                                 & {\color[HTML]{000000} \textbf{0.830}} & \textbf{0.887}                        & 0.785                                 & 0.828                                 \\
change of color saturation           & 0.209   & 0.214   & 0.156  & 0.048                                 & 0.484                                 & 0.752                                 & {\color[HTML]{000000} \textbf{0.759}} & 0.728                                 & \textbf{0.863}                        \\
multiplicative gaussian noise        & 0.688   & 0.777   & 0.583  & 0.731                                 & 0.787                                 & 0.515                                 & {\color[HTML]{000000} \textbf{0.835}} & 0.754                                 & \textbf{0.842}                        \\
comfort noise                        & 0.299   & 0.258   & 0.573  & 0.566                                 & {\color[HTML]{000000} \textbf{0.819}} & 0.773                                 & 0.812                                 & 0.816                                 & \textbf{0.904}                        \\
lossy compression of noisy   images  & 0.677   & 0.573   & 0.717  & 0.727                                 & 0.895                                 & 0.877                                 & \textbf{0.949}                        & 0.830                                 & {\color[HTML]{000000} \textbf{0.933}} \\
color quantization with dither       & 0.775   & 0.782   & 0.699  & 0.802                                 & 0.822                                 & 0.848                                 & \textbf{0.917}                        & 0.702                                 & {\color[HTML]{000000} \textbf{0.856}} \\
chromatic aberrations                & 0.732   & 0.768   & 0.686  & 0.721                                 & 0.762                                 & 0.744                                 & 0.798                                 & \textbf{0.932}                        & {\color[HTML]{000000} \textbf{0.816}} \\
sparse sampling and   reconstruction & 0.848   & 0.857   & 0.888  & 0.856                                 & \textbf{0.917}                        & 0.882                                 & {\color[HTML]{000000} \textbf{0.913}} & 0.769                                 & 0.839                                 \\ \hline
\end{tabular}
\end{table*}

In Table \ref{table:4}, the evaluation of a single distortion type through experiments was carried out using the CSIQ dataset. The comparison was conducted using the same methodological approach as detailed in Table \ref{table:3}. Based on the findings from the experimental results, it is evident that optimal performance is achieved by the proposed method when applied to four specific types of distortion, Gaussian white noise, Jpeg compression, Jpeg2000 compression, additive Gaussian pink noise and Gaussian blur. Nevertheless, poor performance is noted for the method when dealing with contrast-curtailed (CC) distortion. The potential reason for this could be identified as the reduced differentiation between high and low frequency components of the image caused by contrast curtailment.

\begin{table*}
\centering
\caption{Cross-Dataset Performance Comparisons.}
\label{table:6}
\begin{tabular}{c|ccc|ccc|ccc|ccc}
\hline
Train on          & \multicolumn{3}{c|}{LIVE}                                                                                             & \multicolumn{3}{c|}{TID2013}                                                                                          & \multicolumn{3}{c|}{CSIQ}                                                                                             & \multicolumn{3}{c}{LIVEC}                                                                                             \\ \hline
Test on           & TID2013                               & CSIQ                                  & LIVEC                                 & LIVE                                  & CSIQ                                  & LIVEC                                 & LIVE                                  & TID2013                               & LIVEC                                 & LIVE                                  & TID2013                               & CSIQ                                  \\ \hline
DIIVINE\cite{moorthy2011blind}           & 0.342                                 & 0.602                                 & 0.296                                 & 0.714                                 & 0.583                                 & 0.235                                 & 0.817                                 & 0.417                                 & 0.366                                 & 0.354                                 & 0.327                                 & 0.419                                 \\
BRISQUE\cite{mittal2012no}           & 0.354                                 & 0.573                                 & 0.326                                 & 0.724                                 & 0.568                                 & 0.109                                 & 0.823                                 & 0.433                                 & 0.106                                 & 0.244                                 & 0.275                                 & 0.236                                 \\
CORNIA\cite{ye2012unsupervised}            & 0.414                                 & 0.619                                 & 0.429                                 & 0.809                                 & 0.659                                 & {\color[HTML]{000000} \textbf{0.263}} & 0.836                                 & 0.332                                 & {\color[HTML]{000000} \textbf{0.406}} & 0.553                                 & 0.385                                 & 0.433                                 \\
HOSA\cite{xu2016blind}              & 0.47                                  & 0.596                                 & 0.455                                 & 0.844                                 & 0.609                                 & 0.244                                 & 0.786                                 & 0.341                                 & 0.277                                 & 0.522                                 & 0.372                                 & 0.334                                 \\
CNN\cite{kang2014convolutional}               & 0.407                                 & 0.616                                 & 0.103                                 & 0.532                                 & 0.598                                 & 0.104                                 & 0.713                                 & 0.315                                 & 0.013                                 & 0.103                                 & 0.021                                 & 0.095                                 \\
WaDIQaM\cite{bosse2017deep}           & 0.396                                 & 0.601                                 & 0.151                                 & 0.805                                 & 0.683                                 & 0.009                                 & 0.813                                 & 0.506                                 & 0.106                                 & 0.323                                 & 0.141                                 & 0.323                                 \\
TS-CNN\cite{yan2018two}            & 0.431                                 & 0.621                                 & 0.273                                 & 0.566                                 & 0.613                                 & 0.014                                 & 0.848                                 & 0.484                                 & 0.169                                 & 0.081                                 & 0.049                                 & 0.200                                 \\
RAN4IQA\cite{ren2018ran4iqa}           & 0.466                                 & 0.64                                  & 0.134                                 & 0.811                                 & 0.694                                 & 0.036                                 & 0.806                                 & 0.471                                 & 0.116                                 & 0.296                                 & 0.157                                 & 0.276                                 \\
Hall-IQA\cite{lin2018hallucinated}          & 0.486                                 & 0.668                                 & 0.126                                 & 0.786                                 & 0.683                                 & 0.116                                 & 0.833                                 & 0.491                                 & 0.107                                 & /                                     & /                                     & /                                     \\
MMMNet\cite{li2021mmmnet}            & \textbf{0.546}                        & \textbf{0.793}                        & {\color[HTML]{000000} \textbf{0.502}} & {\color[HTML]{000000} \textbf{0.853}} & {\color[HTML]{000000} \textbf{0.702}} & \textbf{0.348}                        & 0.890                                 & 0.522                                 & {\color[HTML]{000000} \textbf{0.406}} & {\color[HTML]{000000} \textbf{0.528}} & 0.398                                 & 0.518                                 \\
VCRNet\cite{pan2022vcrnet}            & {\color[HTML]{000000} \textbf{0.521}} & 0.693                                 & \textbf{0.566}                        & 0.843                                 & 0.660                                 & {\color[HTML]{000000} \textbf{0.263}} & {\color[HTML]{000000} \textbf{0.916}} & {\color[HTML]{000000} \textbf{0.528}} & \textbf{0.422}                        & 0.428                                 & {\color[HTML]{000000} \textbf{0.309}} & {\color[HTML]{000000} \textbf{0.553}} \\
\textbf{proposed} & 0.489                                 & {\color[HTML]{000000} \textbf{0.712}} & 0.472                                 & \textbf{0.863}                        & \textbf{0.731}                        & 0.240                                 & \textbf{0.931}                        & \textbf{0.576}                        & 0.323                                 & \textbf{0.740}                        & \textbf{0.401}                        & \textbf{0.602}                        \\ \hline
\end{tabular}
\end{table*}

Table \ref{table:5} presents the performance metrics of the proposed method as evaluated on the TID2013 dataset. As the most challenging comprehensive distortion IQA dataset, TID2013 contains $24$ distortion types. The comparison includes traditional methods with recent SOTA methods (e.g., MMMNet, VCRNet). Upon examination of the table, it is evident that the method in question has demonstrated exceptional performance across $20$ distinct categories out of a total of $24$ distortion types considered. Among the challenging distortion types, which include impulse noise, quantization noise, and comfort noise, the proposed method is noted to have achieved the highest level of performance when compared to other methods within these specific categories. Specifically, the proposed method outperforms the second best performance in these distortions by $5.1\%$,$ 5.9\%$, and $8.5\%$, respectively.
\subsection{Cross-dataset Experiments}

To demonstrate the generalizability of the proposed method, a comprehensive set of cross-dataset experiments was conducted, including the LIVE, CSIQ, TID2013, and LIVEC datasets. As depicted in Table \ref{table:6}, the training phase is conducted on a single dataset, followed by testing on the three remaining datasets. Results obtained from the prediction models associated with databases that utilize differing scales of subjective image quality are subjected to a scaling process. Post this scaling, the results are aligned and compared on a uniform scale to facilitate a meaningful comparison. For the comparative analysis, a selection of methods was made, consisting of four traditional NR-IQA approaches, namely DIIVINE, BRISQUE, CORNIA, and HOSA, in addition to seven methods that are predicated on deep learning, including CNN, WaDIQaM, TS-CNN, RAN4IQA, Hall-IQA, MMMNet, and VCRNet. Upon reviewing the table, it is evident that out of the entire set of $12$ items subjected to comparison, the proposed method has demonstrated a high level of performance in $7$ items. In particular, training at CSIQ and testing at LIVE and TID2013 achieved sparkling results, outperforming the second best by $1.5\%$ and $4.8\%$, respectively. Due to the guidance of the high-frequency network for the quality estimation network, the proposed method shows strong generalization ability in predicting images with unknown distortion types.

\subsection{Ablation Experiment}
Table \ref{table:7} presents the outcomes of ablation experiments, aimed at demonstrating the distinct contributions and impacts of each component within the proposed model. The high-frequency information embedded within the image is effectively harnessed by the proposed high-frequency extraction network, which serves to enhance the network's attentiveness to the distortion features that hold significant importance for the Human Visual System (HVS). The term "Addition" is utilized to denote the instance where the attentional feature fusion aspect of the proposed model has been substituted by an addition operation. "Vanilla Convolution" refers to the experiment of using vanilla convolution instead of octave convolution in the high frequency extraction stream. Good performance was achieved in both the CSIQ and TID2013 datasets. As evidenced by the table \ref{table:7}, the incorporation of an attention-based feature fusion method has led to an improvement in the network's capacity to represent distorted features. This enhancement is manifested in the form of more effective performance as opposed to those achieved through conventional methods of feature addition, improves $1.0\%$,$1.9\%$ (PLCC, SRCC) and $1.7\%$,$1.5\%$ (PLCC, SRCC) in CSIQ and TID2013 datasets, respectively. The octave convolution employed in this paper is effective in enhancing the model's performance compared to vanilla convolution. This approach has resulted in improvements of $1.0\%$,$1.9\%$ (PLCC, SRCC) and $1.8\%$, $2.0\%$ (PLCC, SRCC) on the CSIQ and TID2013 datasets, respectively.

\begin{table}
\centering
\caption{Performance Comparison of Different Components.}
\label{table:7}
\begin{tabular}{c|cc|cc}
\hline
Dataset                  & \multicolumn{2}{c|}{CSIQ}       & \multicolumn{2}{c}{TID2013}     \\ \hline
Metrics                  & \textbf{PLCC}  & \textbf{SRCC}  & \textbf{PLCC}  & \textbf{SRCC}  \\ \hline
Content-aware Subtask    & 0.945          & 0.931          & 0.903          & 0.883          \\
Distortion-aware Subtask & 0.951          & 0.931          & 0.902          & 0.875          \\
Addition                 & 0.957          & 0.943          & 0.899          & 0.882          \\
Vanilla Convolution      & 0.957          & 0.943          & 0.898          & 0.877          \\
proposed                 & \textbf{0.967} & \textbf{0.962} & \textbf{0.916} & \textbf{0.897} \\ \hline
\end{tabular}
\end{table}

\subsection{gMAD Experiment}

\begin{figure*}[htbp]
\centerline{\includegraphics[width=11cm]{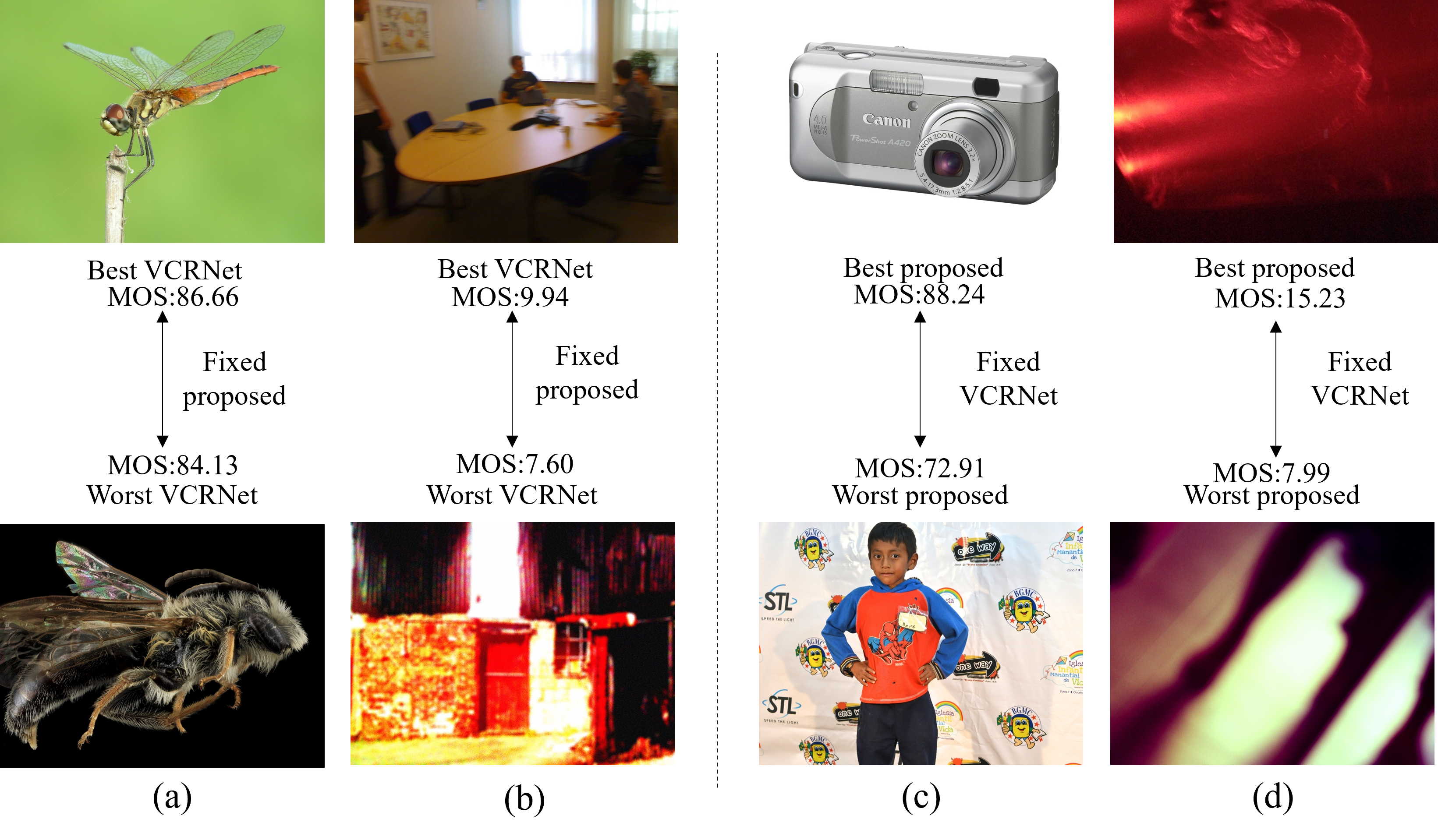}}
\caption{The gMAD competition results between VCRNet and our proposed method. (a) Fixed proposed at the high-quality level. (b) Fixed proposed at the low-quality level. (c) Fixed VCRNet at the high-quality level. (d) Fixed VCRNet at the low-quality level. To better compare the differences between the two images, the ground truth (MOS) of the images has been shown. }
\label{fig:gmad}
\end{figure*}

To further evaluate the performance of the proposed method, the group MAximum Differentiation (gMAD) competition was conducted\cite{ma2016group}. gMAD is based on the idea that an attempt is made to disprove a model, and a model that is more difficult to disprove is considered to be a relatively better model. gMAD utilizes an attacker model to search for counterexamples in the database. These counterexamples are optimal for the attacker model, and if the attack is successful, the defender model is disproved. If, on the contrary, the defender survives such an attack, this indicates that it may be a robust and reliable model. Fig. \ref{fig:gmad} illustrates the results of our experiments. A comparative analysis was performed, positioning it against the current state-of-the-art model known as VCRNet. In the Fig. \ref{fig:gmad}, (a) and (b) denote the image cohorts selected by our proposed model as the defender and VCRNet as the attacker in two sets of images at high and low levels, respectively. (c) and (d) show the cases where our proposed model acts as the attacker and VCRNet acts as the defender. As depicted in the Fig. \ref{fig:gmad}, it can be observed that the image selected by the attacker undergoes minimal perceptual change when the proposed method is operational in the capacity of a defender. On the contrary, when the proposed method functions in the capacity of an attacker, image pairs with significantly larger differences are efficiently recognized and selected. This reflects the robustness of our method.
\section{Conclusion}
In this paper, we present an end-to-end multi-task no-reference image quality assessment framework. The proposed framework is distinguished by its innovative three-branch architecture, comprising a high-frequency extraction network, a quality estimation network, and a distortion-aware network. The high-frequency extraction network, by capitalizing on the human visual system's sensitivity to high-frequency image components, effectively guides the quality estimation network to focus on perceptually significant regions. The distortion-aware network is pre-trained using a contrastive learning method, which enhances the generalization ability to authentic distortion. Furthermore, a feature fusion module (FFM) is proposed to enhance the capability of image quality prediction by effectively fusing features from multiple networks through the adoption of a feature fusion method based on the attention mechanism. The evaluations on five benchmark IQA datasets have conclusively demonstrated the superior performance of our NR-IQA framework. The results not only affirm the model's state-of-the-art accuracy but also highlight its robustness across various image distortions.

\bibliographystyle{IEEEtran}
\bibliography{paper}


 




\vfill

\end{document}